\documentclass[conference, letterpaper]{IEEEtran}
\IEEEoverridecommandlockouts
\usepackage[table]{xcolor}
\usepackage{geometry}
\usepackage[colorlinks,
      linkcolor=blue,
      anchorcolor=blue,
      citecolor=blue,
      urlcolor=blue]{hyperref} 
\usepackage{cite}
\usepackage{amsmath,amssymb,amsfonts}
\usepackage{algorithmic}
\usepackage{subcaption}
\usepackage{algorithm}
\usepackage{graphicx}
\usepackage{float}
\usepackage{diagbox}
\usepackage{soul}
\usepackage{soul}
\usepackage{makecell}
\usepackage[font=footnotesize]{caption}
\usepackage{textcomp}
\usepackage{siunitx}   
\usepackage{booktabs}
\usepackage{soul}
\usepackage{booktabs}        
\usepackage{threeparttable}  
\usepackage{amssymb}         
\usepackage{tikz}
\usetikzlibrary{shapes, arrows.meta, positioning, fit, backgrounds, calc, shadows}
\definecolor{predColor}{RGB}{230, 240, 250} 
\definecolor{planColor}{RGB}{255, 250, 230} 
\definecolor{plantColor}{RGB}{235, 250, 235} 
\definecolor{ctrlColor}{RGB}{240, 240, 240} 

\def\BibTeX{{\rm B\kern-.05em{\sc i\kern-.025em b}\kern-.08em
  T\kern-.1667em\lower.7ex\hbox{E}\kern-.125emX}}
  
\begin{document}
\title{\LARGE \bf MoE-ACT: Scaling Multi-Task Bimanual Manipulation with Sparse Task-Conditioned Mixture-of-Experts Transformers}
\author{Kangjun Guo, Haichao Liu, Yanji Sun, Ruhan Zhao, Jinni Zhou, Jun Ma, \textit{Senior Member, IEEE} \thanks{Kangjun Guo, Haichao Liu, Yanji Sun, Ruhan Zhao, Jinni Zhou are with The Hong Kong University of Science and Technology (Guangzhou), Guangzhou 511453, China (e-mail: \{kguo886, hliu369, ysun993, rzhao238\}@connect.hkust-gz.edu.cn; eejinni@hkust-gz.edu.cn).}
\thanks{Jun Ma is with The Hong Kong University of Science and Technology (Guangzhou), Guangzhou 511453, China, and also with The Hong Kong University of Science and Technology, Hong Kong SAR, China (e-mail: jun.ma@ust.hk).}
}

\maketitle

\begin{abstract}
Developing a unified policy for multi-task robotic manipulation remains challenging due to policy degradation from task interference and negative transfer. In this work, we propose Mixture-of-Experts-Enhanced Action Chunking Transformer (MoE-ACT), a parameter-efficient multi-task visuomotor framework tailored for bimanual manipulation. MoE-ACT incorporates sparse MoE layers into the ACT encoder, dynamically routing image tokens to selected experts based on task context, visual observations, and proprioceptive states. Furthermore, the framework incorporates task-conditioned Feature-wise Linear Modulation (FiLM) in the action decoder alongside multi-scale cross-attention, ensuring precise task grounding and capturing fine-grained spatial cues. Extensive evaluations on the RoboTwin 2.0 benchmark across 16 challenging bimanual tasks demonstrate that MoE-ACT achieves an average success rate of 62.0\%, outperforming standard ACT by 17.4 percentage points. Crucially, with only 195M activated parameters, MoE-ACT surpasses the 16-fold larger foundation model $\pi_0$ (3.24B) by 5.1 percentage points, exhibiting substantial gains in parameter efficiency and deployment feasibility. Real-robot dual-arm experiments consistently confirm its superior multi-task execution capabilities. Our open-source project page can be found at \url{https://j3k7.github.io/MoE-ACT/}.
\end{abstract}

\begin{figure*}[t]
\centering
\includegraphics[width=1.0\linewidth]{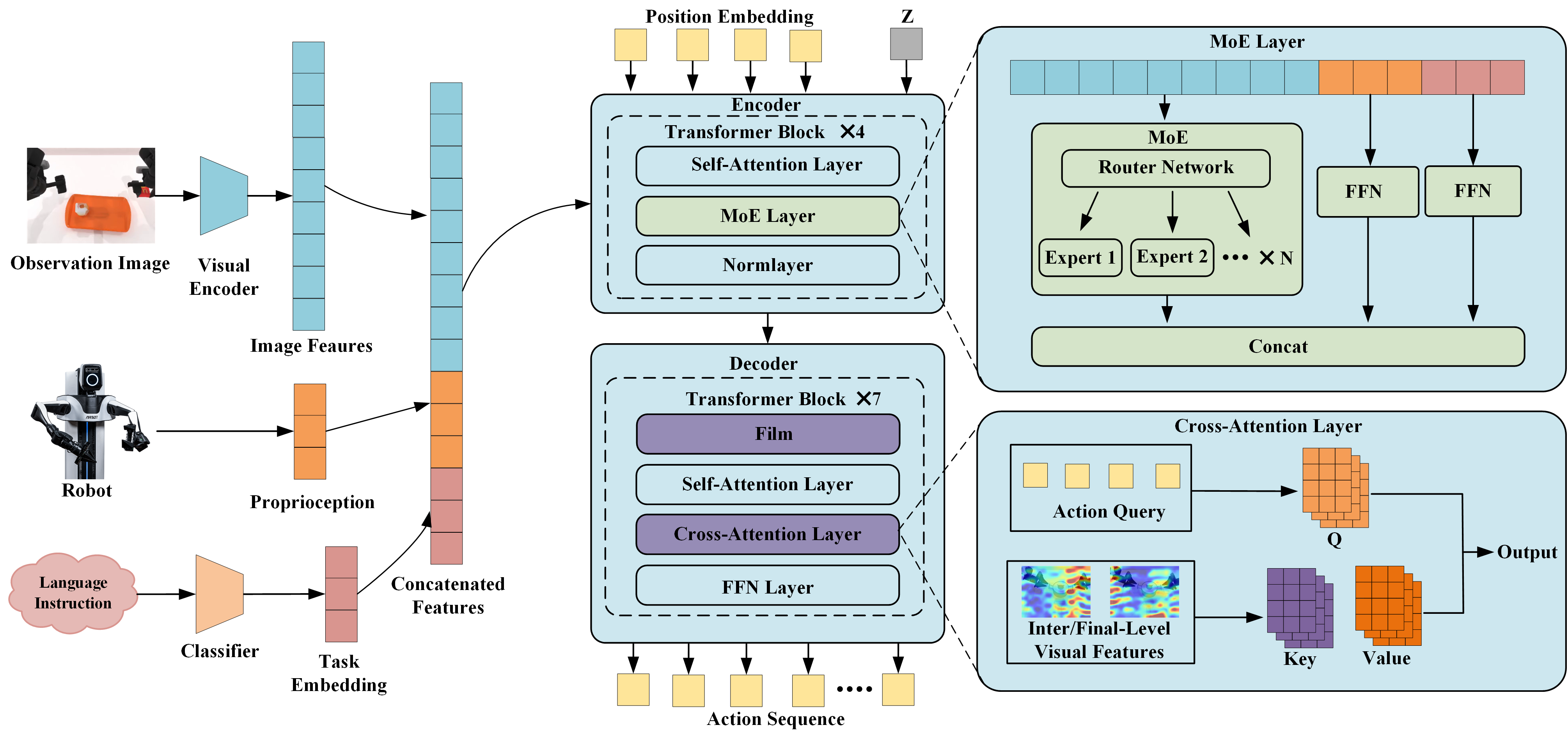}
\vspace{-1.7em}
\caption{Overview of MoE-ACT. The architecture consists of the MoE module integrated into the Transformer encoder and a FiLM mechanism in the decoder. The MoE module enables task-specific feature decoupling, while FiLM ensures that action generation is consistent with task instructions. Multi-scale cross-attention allows the model to capture both high-level semantics and low-level visual details for manipulation control.}
\label{fig:4}
\vspace{-0.3cm}
\end{figure*}

\section{Introduction}

Robotic manipulation is important in both industrial and household settings, where it can reduce human labor requirements. Recent advances in imitation learning algorithms, including Diffusion Policy\cite{1} and Action Chunking with Transformers (ACT)\cite{2}, have enabled robots to perform individual manipulation tasks with high success rates. However, practical deployment often requires a single robot to perform a variety of tasks within a unified policy\cite{3,29}. This motivates the development of effective multi-task learning methods for robotic manipulation.

Accurate modeling of action distributions is central to learning robotic manipulation policies. Robotic actions are typically represented in high-dimensional continuous spaces, where different action sequences may accomplish the same task, resulting in multimodal action distributions. Diffusion policies have achieved strong performance on complex manipulation tasks by formulating action generation as an iterative denoising process\cite{4,5}. ACT uses a CVAE to model variations in expert demonstrations and predicts action sequences in chunks to mitigate compounding errors in imitation learning. However, in multi-task settings, conflicting task objectives and overlapping action distributions can interfere with learning task-specific behaviors\cite{6,7}. When multiple tasks share the same policy parameters, learning one task may adversely affect performance on others. Such interference limits the policy’s ability to accommodate task-specific behaviors.

Large-scale Vision-Language-Action (VLA) models have emerged as a promising paradigm for general-purpose robotic policies\cite{8,9,10,44,11}. Representative examples, such as $\pi_0$\cite{10}, \mbox{$\pi_{0.5}$}\cite{44}, and RDT\cite{11}, typically contain billions of parameters. Pre-training on large multimodal datasets substantially improves zero-shot generalization and open-world task execution. However, training these models requires substantial computational resources and large amounts of data. Their large model sizes can also increase inference latency, posing challenges for real-time robotic control\cite{12}. These considerations motivate the development of multi-task policies that balance manipulation performance with computational efficiency. Our work focuses on learning a unified policy for multiple bimanual manipulation tasks to improve performance while maintaining a relatively low parameter count.

A promising architecture for improving multi-task performance while maintaining computational efficiency is MoE framework\cite{14,15}. MoE has been widely adopted in large language models\cite{16,17} and multi-task robotics\cite{25,26,27}. It decomposes a model into specialized expert components, each focused on different tasks or data characteristics. The model then selectively activates relevant experts based on the input, preserving computational efficiency.

In this work, we propose MoE-ACT, a unified policy for multi-task bimanual manipulation. We integrate the MoE module into the Transformer encoder. The router combines task context with visual and proprioceptive information to select experts for processing image tokens. We also incorporate learnable task embeddings to provide task information and guide action generation. In the decoder, task-conditioned FiLM modulates action features\cite{38,39}, while multi-scale cross-attention provides visual information from different processing stages\cite{40}. By combining sparse expert routing with task conditioning and multi-scale visual fusion, MoE-ACT improves the ability of a unified policy to handle diverse bimanual manipulation tasks.

In summary, the contributions of this paper are as follows:
\begin{itemize}

\item We propose MoE-ACT, a unified policy for multi-task bimanual manipulation that integrates MoE modules into ACT. By selectively activating different experts, the policy learns task-specific representations and reduces interference during joint training.

\item Within MoE-ACT, we design a task-conditioned action decoder that combines FiLM conditioning with multi-scale cross-attention. These complementary mechanisms integrate task semantics and multi-scale visual information to guide action generation.

\item The effectiveness of the proposed method is validated through both simulation and real-world experiments. Ablation studies further evaluate the effects of the number of experts, FiLM conditioning, and multi-scale cross-attention on multi-task manipulation performance.
\end{itemize}

\section{Related Works}
\subsection{Imitation Learning for Robotic Manipulation}
Recently, imitation learning has made significant progress in helping robots acquire complex manipulation skills\cite{1, 2, 22}. This paradigm learns directly from expert demonstration data, allowing robots to efficiently capture complex trajectory distributions and establish a direct mapping from perception to action\cite{30}.

As a representative architecture in this field, the ACT policy\cite{2} is trained via a CVAE and models the action prediction as action chunks to reduce compounding errors. Compared to diffusion policies, ACT provides faster inference as it avoids the time-consuming iterative denoising process. However, due to the lack of language modality input, the original ACT is unable to leverage semantic context to focus on task-relevant visual features. To address this issue, MT-ACT\cite{24} introduces task language instructions to enhance the multi-task learning capabilities of ACT. Additionally, BAKU\cite{23} combines a Transformer architecture with action chunking for efficient multi-task robot learning. Its visual encoder uses FiLM to incorporate task information during visual feature extraction. Despite these advances, multi-task learning remains a significant challenge for imitation learning.

With the advancement of multimodal learning, VLA models have become a major approach to general-purpose robot control\cite{32}. Models such as RT-2\cite{8} and OpenVLA\cite{9} connect pretrained vision-language representations with robot actions, enabling semantic reasoning and generalization across diverse tasks. In parallel, world models predict how the environment evolves under candidate actions, allowing robots to plan through predicted future states. For example, DINO-WM learns visual dynamics for zero-shot planning\cite{41}, while Cosmos Policy adapts a pretrained video model to generate robot actions and future states\cite{42}. More recently, World Action Models (WAMs), such as DreamZero, jointly model future visual observations and robot actions to improve physical generalization\cite{43}. However, large-scale VLA and video-world-model policies still require extensive pretraining data and substantial computation, resulting in high training costs, memory consumption, and inference latency\cite{33,42,43}.

To address these challenges, this work introduces MoE-ACT that incorporates language instruction as a semantic guide. And MoE-ACT integrates the MoE mechanism into the Transformer encoder to decouple task representations and enhance multi-task performance.

\subsection{Multi-task Learning in Robotics}
Multi-task learning enables robots to perform varied tasks in diverse settings. Nowadays, significant advancements have been achieved in multi-task learning in robotic manipulation. Training on large-scale datasets, such as Open X-Embodiment\cite{34} and DROID\cite{35}, can significantly enhance the generalization performance and multi-task capabilities of robot models in complex scenarios. For example, $\pi_0$\cite{10} demonstrates superior performance in handling complex manipulation tasks through joint training on large-scale mixed robot datasets.

Many studies have advanced multi-task learning through innovations in policy architecture\cite{37,38,7,36}. For instance, skill-based policies significantly enhance model generalization by abstracting and reusing task-invariant skills across various tasks. Discrete Policy\cite{7} employs VQ-VAE to abstract continuous action sequences into discrete latent skill identifiers. By integrating generative models within a decoupled latent space, this approach enables the reuse of various manipulation skills across different tasks. Similarly, SkillDiffuser\cite{36} combines discrete skill representations with conditional diffusion models to map abstract instructions to consistent action sequences, significantly improving skill reuse and model generalization in multi-task scenarios.

The MoE architecture has shown strong potential for multi-task robotic learning. By enabling different subnetworks to specialize in distinct skills, MoE improves task decomposition and overall robustness. In robotic manipulation, MoE has been applied to improve diffusion policies\cite{18,20,21}. For example, MoE-DP inserts an MoE layer between the vision encoder and the diffusion model, improving robustness to interference and long-horizon task execution\cite{20}. MoE is also used to scale model capacity and regulate computational flow across diffusion-policy modules\cite{21}. MoE has demonstrated its effectiveness in Transformer-based large language models\cite{16}, motivating its integration into the Transformer-based ACT policy. Integrating MoE with ACT preserves high inference efficiency while offering substantial performance gains in multi-task settings.

\section{Methodology}
\label{sec:method}
MoE-ACT is a unified framework designed for robotic multi-task manipulation. Fig.~\ref{fig:4} shows the overview of MoE-ACT. The approach builds upon ACT\cite{2}. The integration of the MoE mechanism within the Transformer encoder and a FiLM mechanism within the decoder can address the challenges of negative transfer and task interference in multi-task learning. Furthermore, an additional cross-attention layer is introduced in the decoder to enhance the model's perception of low-level visual features.

\subsection{Problem Formulation and Overview}
This work considers a multi-task imitation learning setting in which a robot learns a policy $\pi$ to perform $N_{task}$ distinct tasks. At each time step $t$, the policy receives an observation $\mathcal{O}_t$ consisting of joint positions $q_t \in \mathbb{R}^{d_q}$ and RGB images $I_t \in \mathbb{R}^{H \times W \times 3}$ from multiple camera views. Each task instruction \mbox{$l$} is associated with a learnable task embedding \mbox{$z_{task}$} that conditions both expert routing and action decoding. The objective is to predict a sequence of future actions \mbox{$\mathbf{a}_{t:t+T_a-1}$}, where \mbox{$T_a$} denotes the action chunk size.
The overall MoE-ACT architecture contains three main components: (1) an MoE-enhanced encoder that processes visual and proprioceptive inputs and dynamically routes image tokens to specific experts to decouple feature representations; (2) a Transformer decoder that generates actions conditioned on the task embedding via FiLM and attends to multi-scale features to improve manipulation performance; and (3) a CVAE encoder that maps action sequences into a latent space $z$ to implicitly capture diverse behavioral modes in the data.

\subsection{MoE-Enhanced Encoder}

Standard Transformers process all inputs with shared parameters, which can cause conflicting gradient updates across diverse tasks. To mitigate this issue, we integrate an MoE module into the feed-forward networks (FFNs) of the Transformer encoder layers. The MoE module consists of a routing network and a collection of $N$ expert networks (FFNs). The routing network dynamically computes the activation weight of each expert from the input features. We adopt sample-level routing, where all image tokens from the same multi-view observation share the selected experts and routing weights. By incorporating both task information and proprioceptive states, the router selects experts based on the task requirements and the robot's joint positions, in addition to the visual observations. The selected expert subset can therefore vary across observations, allowing different stages of the same task to use different experts. Let the input sequence to the \mbox{$\ell$}-th encoder layer be $X^\ell = [x_{latent}, x_{prop}, x_{task}, x_{img}^1, \dots, x_{img}^M]$, where \mbox{$x_{latent}$} denotes the projected CVAE latent token. The gating input \mbox{$h_{\text{router\_input}}$} is constructed by concatenating proprioception, the task embedding, and the global average of image tokens:
\begin{equation}
    h_\text{router\_input} = \text{Concat}\left(x_{prop}, \; x_{task}, \; \frac{1}{M}\sum_{j=1}^{M} x_{img}^j\right).
\end{equation}
The gating network integrates the concatenated features to produce task-conditioned routing logits:
\begin{equation}
    H(x) = \displaystyle G_{\theta}(h_{\text{router\_input}})+ \epsilon, \quad \epsilon \sim \mathcal{N}(0, \sigma^2),
\end{equation}
where \mbox{$G_{\theta}$} denotes the learnable gating network, \mbox{$N$} is the total number of experts, and $\epsilon$ is Gaussian noise added during training for exploration.
The router selectively activates the top-$k$ experts with the highest logits.
The gating weights are obtained via Softmax over the selected experts:

\begin{equation}
    P(e|x)_i = 
    \begin{cases}
    \frac{\exp(H(x)_i)}{\sum_{j \in \text{Top-}k} \exp(H(x)_j)} & \text{if } i \in \text{Top-}k(H(x)) \\
    0 & \text{otherwise}
    \end{cases}.
\end{equation}
The final MoE output is the probability-weighted sum of the activated expert outputs. This sparse activation mechanism increases model capacity while limiting computation to the selected experts for each input. Non-image tokens are processed by dedicated FFNs, while image tokens are routed to the selected experts. For image tokens $x_{img}$, the output is the weighted sum of selected experts:
\begin{equation}
    \text{FFN}_{MoE}(x_{img}) = \sum_{i=1}^{N} P(e|x)_i \cdot E_i(x_{img}),
\end{equation}
where $E_i(\cdot)$ represents the $i$-th expert network.

\subsection{Task-Conditioned Transformer Decoder}
The decoder is responsible for translating the encoded features into precise action sequences. We enhance the standard Transformer decoder with two mechanisms: task conditioning via FiLM and multi-scale feature fusion.

\subsubsection{Task-conditioned FiLM}
At each decoder layer, FiLM uses \mbox{$z_{task}$} to generate task-specific scale (\mbox{$\gamma$}) and shift (\mbox{$\beta$}) parameters:
\begin{equation}
    [\gamma, \beta] = \text{MLP}_{FiLM}(z_{task}).
\end{equation}
For the action query \mbox{$Q \in \mathbb{R}^{T_a \times d_{model}}$}, the modulated query $Q'$ is formulated as:
\begin{equation}
    Q' = \text{FiLM}(Q, z_{task}) = (1 + \gamma) \odot Q + \beta,
\end{equation}
where $\odot$ denotes element-wise multiplication. This modulation incorporates task information into the action queries, helping align the generated actions with the intended task.

\subsubsection{Multi-Scale Cross-Attention Layer}
Standard decoders attend only to the final encoder output, which can discard low-level visual details. Additional cross-attention in the first five decoder layers integrates features from the visual backbone and different encoder layers. These layers attend to the projected backbone output and the four encoder-layer outputs, respectively, before attending to the final encoder representation. All decoder layers retain cross-attention to the final encoder representation. For each of the first five layers, the two cross-attention blocks are expressed as:
\begin{align}
    \text{Out}_{inter}^{(d)} &= \displaystyle \text{CA}(\widetilde{Q}^{(d)}, F_{inter}^{(d)}), \\
    \text{Out}_{final}^{(d)} &= \displaystyle \text{CA}(\text{Out}_{inter}^{(d)}, F_{enc}).
\end{align}
Here, \mbox{$\text{CA}(Q,F)$} denotes a cross-attention block with action features \mbox{$Q$} as queries and features \mbox{$F$} as keys and values. The action features \mbox{$\widetilde{Q}^{(d)}$} are obtained after FiLM modulation and self-attention, and \mbox{$F_{inter}^{(d)}$} denotes the features assigned to decoder layer \mbox{$d$}. $F_{enc}$ is the final output of the MoE encoder, containing high-level contextual information. This design combines visual information from different processing stages to support action prediction.

\subsection{Training Objectives}
The model is trained end-to-end to minimize a composite loss function $\mathcal{L}_{total}$:
\begin{equation}
    \mathcal{L}_{total} = \mathcal{L}_{rec} + \lambda_{kl} \mathcal{L}_{kl} + \lambda_{MoE} \mathcal{L}_{MoE},
\end{equation}
where $\mathcal{L}_{rec}$ is the action reconstruction loss for imitation learning, $\mathcal{L}_{kl}$ represents the Kullback-Leibler divergence loss for regularizing the CVAE latent space, and $\mathcal{L}_{MoE}$ is the auxiliary loss for the MoE module. The coefficients $\lambda_{kl}$ and $\lambda_{MoE}$ are hyperparameters that weigh the contribution of the regularization and auxiliary terms, respectively.

\subsubsection{CVAE Loss}
The CVAE encoder models the latent posterior conditioned on joint states and demonstrated actions. During training, sampled \mbox{$z$} is projected into a token for the MoE encoder; at inference, \mbox{$z=0$}. Following ACT, training combines L1 action reconstruction with KL regularization:
\begin{equation}
\begin{split}
    \mathcal{L}_{rec} &= \sum_{t} \| \hat{\mathbf{a}}_t - \mathbf{a}_t \|_1, \\
    \mathcal{L}_{kl} &= D_{KL}\left(q(z|q_t, \mathbf{a}_{t:t+T_a-1}) \| p(z)\right),
\end{split}
\end{equation}
where \mbox{$p(z)=\mathcal{N}(0,I)$} is the latent prior.

\subsubsection{MoE Auxiliary Loss}
To prevent mode collapse and ensure expert diversity, we introduce three auxiliary losses for the MoE module:
\begin{equation}
    \displaystyle \mathcal{L}_{MoE} = -\lambda_{ent} \mathcal{L}_{entropy} + \lambda_{bal} \mathcal{L}_{balance} + \lambda_{orth} \mathcal{L}_{orth}.
\end{equation}
\begin{itemize}
    \item \textbf{Entropy Loss} ($\mathcal{L}_{entropy}$): Maximizes the entropy of the routing distribution to encourage the exploration of all experts.
    \item \textbf{Load Balancing Loss} ($\mathcal{L}_{balance}$): Penalizes the variance in expert assignment frequency to ensure an equal computational load across experts.
    \item \textbf{Orthogonal Loss} ($\mathcal{L}_{orth}$): Minimizes the cosine similarity between expert weight matrices, encouraging experts to learn distinct features.
\end{itemize}

\section{Experimental Results and Analysis}
\label{sec:experiments}

To evaluate the proposed method, we conduct multi-task learning experiments on the RoboTwin 2.0 benchmark\cite{28} and in real-world settings. This section includes three parts: (1) comparison between MoE-ACT and representative baselines; (2) ablation studies on key architectural components; and (3) bimanual manipulation experiments on a physical robot to validate the effectiveness of the proposed method in real-world multi-task settings.

\subsection{Simulation Results}
\subsubsection{Environmental Setup and Baselines}
In the multi-task setting, we jointly train a unified policy on 16 representative bimanual manipulation tasks and evaluate it on each task under the Easy (clean) setting of RoboTwin 2.0. MoE-ACT is trained with the Adam optimizer for 1500 epochs, using an initial learning rate of $1 \times 10^{-5}$ and a batch size of 16. In the experiments, the action chunk size is set to 50. Training is performed on two NVIDIA A800 GPUs. The training set contains 800 episodes in total, with 50 episodes per simulation task. For the simulation experiments, MoE-ACT is configured with 64 experts and uses Top-4 expert selection.
We use Diffusion Policy (DP)\mbox{\cite{1}}, Action Chunking Transformer (ACT)\mbox{\cite{2}}, Robotics Diffusion Transformer (RDT)\mbox{\cite{11}}, Transformer-based multi-task policy BAKU\mbox{\cite{23}}, and \mbox{$\pi_0$}\mbox{\cite{10}} as baselines. In the multi-task setting, the ACT baseline is also conditioned on task embeddings to provide explicit task information. The hyperparameters and training settings follow their original implementations. For evaluation, we conduct 50 trials per task and report the average success rates.

\begin{table*}[!t]
\centering
\caption{Bimanual multi-task learning performance on RoboTwin 2.0. Success rates (\%) are reported over 50 evaluation trials per task.}
\label{tab:results}
\setlength{\tabcolsep}{4pt}
\renewcommand{\arraystretch}{1.15}

\begin{tabular}{
l|
*{5}{>{\centering\arraybackslash}p{1.55cm}}
>{\centering\arraybackslash}p{2.45cm}
}
\toprule
\diagbox{\textbf{Task}}{\textbf{Policy (Params.)}}
& \makecell[c]{\textbf{ACT\cite{2}}\\{\scriptsize \textbf{(84M)}}}
& \makecell[c]{\textbf{DP\cite{1}}\\{\scriptsize \textbf{(97M)}}}
& \makecell[c]{\textbf{RDT\cite{11}}\\{\scriptsize \textbf{(1.23B)}}}
& \makecell[c]{\textbf{BAKU\cite{23}}\\{\scriptsize \textbf{(9M)}}}
& \makecell[c]{\textbf{$\boldsymbol{\pi}_0$\cite{10}}\\{\scriptsize \textbf{(3.24B)}}}
& \makecell[c]{\textbf{MoE-ACT (Ours)}\\{\scriptsize \textbf{(983M / 195M active)}}} \\
\midrule

Adjust Bottle          & \textbf{100.0} & 94.0 & 92.0 & \textbf{100.0} & 94.0 & \textbf{100.0} \\
Beat Block Hammer      & 66.0  & 56.0 & 58.0 & 10.0  & 74.0 & \textbf{84.0} \\
Click Alarmclock       & 54.0  & \textbf{86.0} & 82.0 & 62.0  & 52.0 & 58.0 \\
Handover Block         & 0.0   & 26.0 & \textbf{44.0} & 0.0   & 30.0 & 26.0 \\
Lift Pot               & 76.0  & 62.0 & 68.0 & \textbf{78.0} & 28.0 & 76.0 \\
Open Laptop            & 78.0  & 48.0 & 58.0 & 72.0  & 78.0 & \textbf{88.0} \\
Open Microwave         & 10.0  & 52.0 & 46.0 & \textbf{80.0} & 48.0 & 40.0 \\
Pick Dual Bottles      & 36.0  & \textbf{62.0} & 26.0 & 6.0   & \textbf{62.0} & 60.0 \\
Place Bread Skillet    & 0.0   & 18.0 & 20.0 & 6.0   & \textbf{26.0} & 10.0 \\
Place Burger Fries     & 0.0   & 84.0 & 40.0 & 34.0  & 74.0 & \textbf{90.0} \\
Place Container Plate  & 90.0  & 30.0 & 78.0 & 86.0  & \textbf{92.0} & \textbf{92.0} \\
Place Empty Cup        & 70.0  & 42.0 & 44.0 & 28.0  & 66.0 & \textbf{82.0} \\
Place Object Basket    & 32.0  & 30.0 & 32.0 & 42.0  & 42.0 & \textbf{56.0} \\
Put Object Cabinet     & 16.0  & 28.0 & 18.0 & 16.0  & 24.0 & \textbf{30.0} \\
Stack Bowls Two        & 74.0  & 22.0 & 76.0 & \textbf{92.0} & \textbf{92.0} & 88.0 \\
Stamp Seal             & 12.0  & 0.0  & 2.0  & 8.0   & \textbf{28.0} & 12.0 \\

\midrule
\textbf{Average}
& 44.6
& 46.3
& 49.0
& 45.0
& 56.9
& \textbf{62.0} \\

\bottomrule
\end{tabular}
\end{table*}

\subsubsection{Performance and Baseline Comparison}

As shown in Table~\ref{tab:results}, ACT and DP just achieve average success rates of 44.6\% and 46.3\%, respectively. This may be attributed to the high degrees of freedom in bimanual manipulation and the complex multimodal action distributions across multiple tasks. RDT achieves a relatively higher average success rate of 49.0\%, with approximately 1.23B parameters. In contrast, MoE-ACT achieves an average success rate of 62.0\%. It outperforms ACT, DP, RDT, and BAKU by 17.4, 15.7, 13.0, and 17.0 percentage points, respectively. Compared with $\pi_0$, the strongest baseline in terms of average performance, MoE-ACT improves the average success rate by 5.1 percentage points. MoE-ACT contains 983M total parameters, while only 195M parameters are activated during inference. In comparison, $\pi_0$ contains 3.24B parameters. Moreover, MoE-ACT achieves the best or tied-best performance among all evaluated methods on 8 of the 16 tasks. These results demonstrate that sparse expert routing improves multi-task manipulation performance while maintaining a relatively low active parameter count.

\subsection{Ablation Study}

To isolate the contribution of each design, we compare the full MoE-ACT with four variants that remove individual components or use dense parameter scaling:

(1) MoE-ACT: Uses the complete model with the MoE-enhanced encoder, FiLM, and Multi-Scale Cross-Attention.
(2) w/o Multi-Scale Cross-Attention: Removes cross-attention over intermediate visual features.
(3) w/o FiLM: Removes FiLM conditioning from the decoder.
(4) w/o FiLM \& Multi-Scale Cross-Attention: Removes both FiLM conditioning and Multi-Scale Cross-Attention while retaining the MoE-enhanced encoder.
(5) Scaled ACT (0.9B): Scales the original ACT to approximately 0.9B parameters to match the model size of MoE-ACT, without using MoE.

\begin{table*}[!t]
\centering
\begin{threeparttable}
\caption{Ablation results for key components of MoE-ACT on 16 RoboTwin 2.0 tasks.}
\label{tab:1}
\setlength{\tabcolsep}{6pt}
\renewcommand{\arraystretch}{1.15}

\begin{tabular}{l|ccccc}
\toprule
\textbf{Task}
& \textbf{Scaled ACT (0.9B)}
& \textbf{w/o FiLM \& MSCA}
& \textbf{w/o FiLM}
& \textbf{w/o MSCA}
& \textbf{MoE-ACT} \\
\midrule

Adjust Bottle          & \textbf{100.0} & 96.0 & \textbf{100.0} & \textbf{100.0} & \textbf{100.0} \\
Beat Block Hammer      & 52.0  & 52.0 & 68.0  & \textbf{86.0}  & 84.0  \\
Click Alarmclock       & \textbf{76.0}  & 44.0 & 58.0  & 42.0  & 58.0  \\
Handover Block         & 0.0   & 42.0 & 8.0   & \textbf{50.0}  & 26.0  \\
Lift Pot               & 72.0  & 78.0 & \textbf{84.0}  & 78.0  & 76.0  \\
Open Laptop            & \textbf{90.0}  & 82.0 & 82.0  & 80.0  & 88.0  \\
Open Microwave         & 12.0  & 18.0 & 2.0   & \textbf{44.0}  & 40.0  \\
Pick Dual Bottles      & 36.0  & 24.0 & 20.0  & 32.0  & \textbf{60.0}  \\
Place Bread Skillet    & 6.0   & 4.0  & 4.0   & 2.0   & \textbf{10.0}  \\
Place Burger Fries     & 44.0  & 76.0 & 78.0  & 86.0  & \textbf{90.0}  \\
Place Container Plate  & 82.0  & 70.0 & \textbf{94.0}  & \textbf{94.0}  & 92.0  \\
Place Empty Cup        & \textbf{84.0}  & 66.0 & 82.0  & 72.0  & 82.0  \\
Place Object Basket    & 26.0  & 44.0 & 48.0  & 48.0  & \textbf{56.0}  \\
Put Object Cabinet     & 24.0  & 10.0 & 20.0  & 28.0  & \textbf{30.0}  \\
Stack Bowls Two        & 66.0  & 86.0 & \textbf{90.0}  & \textbf{90.0}  & 88.0  \\
Stamp Seal             & 6.0   & 10.0 & \textbf{12.0}  & 8.0   & \textbf{12.0}  \\

\midrule
\textbf{Average}
& 48.5
& 50.1
& 53.1
& 58.8
& \textbf{62.0} \\

\bottomrule
\end{tabular}
\begin{tablenotes}[flushleft]
\footnotesize
\item[] \textit{Note:} Success rates (\%) are reported over 50 evaluation trials per task. MSCA denotes multi-scale cross-attention. Scaled ACT (0.9B) increases the parameter count of ACT without introducing MoE. The ablation variants remove the indicated decoder components while retaining the MoE-enhanced encoder. MoE-ACT includes the MoE-enhanced encoder, FiLM, and MSCA.
\end{tablenotes}
\end{threeparttable}
\end{table*}

\subsubsection{Overall Ablation Results}

As shown in Table~\ref{tab:1}, the full MoE-ACT achieves the highest average success rate of 62.0\%. Performance consistently decreases when the decoder-side components are removed. When both FiLM and Multi-Scale Cross-Attention are removed, the average success rate drops to 50.1\%. This indicates that the two components provide complementary benefits for multi-task learning. Scaled ACT achieves only 48.5\% despite having a comparable total parameter count. This comparison suggests that the improvement of MoE-ACT is not simply due to increased model capacity.

\subsubsection{\texorpdfstring{Effects of FiLM Conditioning and Multi-Scale Cross-Attention}{Effects of FiLM Conditioning and Multi-Scale Cross-Attention}}

The ablation results in Table~\ref{tab:1} show that both FiLM conditioning and Multi-Scale Cross-Attention contribute to the performance of MoE-ACT. Removing FiLM decreases the average success rate from 62.0\% to 53.1\%, corresponding to a drop of 8.9 percentage points, whereas removing Multi-Scale Cross-Attention reduces it to 58.8\%, a drop of 3.2 percentage points. FiLM provides task-dependent modulation for action generation, while Multi-Scale Cross-Attention supplements the final encoder representation with intermediate visual details. When both components are removed, the average success rate further decreases to 50.1\%, indicating that they provide complementary benefits for multi-task manipulation.

\subsubsection{\texorpdfstring{Effect of the Number of Experts}{Effect of the Number of Experts}}

\begin{figure}[!t]
\centering
\includegraphics[width=\linewidth]{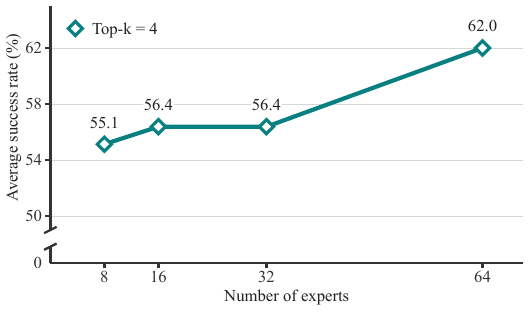}
\caption{Effect of the number of experts on the average success rate over 16 RoboTwin 2.0 tasks.}
\label{fig:expert_scaling}
\end{figure}

To investigate the effect of expert capacity, we evaluate MoE-ACT with 8, 16, 32, and 64 experts on the same 16-task benchmark. As shown in Fig.~\ref{fig:expert_scaling}, the average success rate generally improves as the number of experts increases, rising from 55.1\% with 8 experts to 56.4\% with both 16 and 32 experts, and 62.0\% with 64 experts. Scaling from 8 to 64 experts yields an absolute improvement of 6.9 percentage points. This trend suggests that a larger expert pool provides greater capacity for task-specific specialization and helps reduce interference among heterogeneous manipulation tasks. At the same time, activating only a subset of experts for each input can reduce inference costs to some extent compared with activating all experts.

\subsection{Real-World Experimental Results}
\subsubsection{Experimental Setup}

\begin{figure}[H]
    \centering
    \begin{subfigure}{0.48\linewidth}
        \centering
        \includegraphics[width=\linewidth]{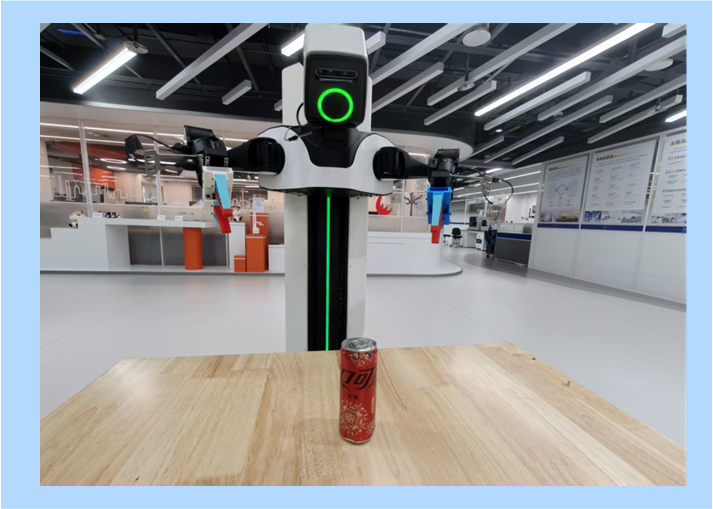}
        \vspace{-1.7em}
        \caption{``Hand over the bottle'' }
        \label{fig:41}
    \end{subfigure}
    \hfill
    \begin{subfigure}{0.48\linewidth}
        \centering
        \includegraphics[width=\linewidth]{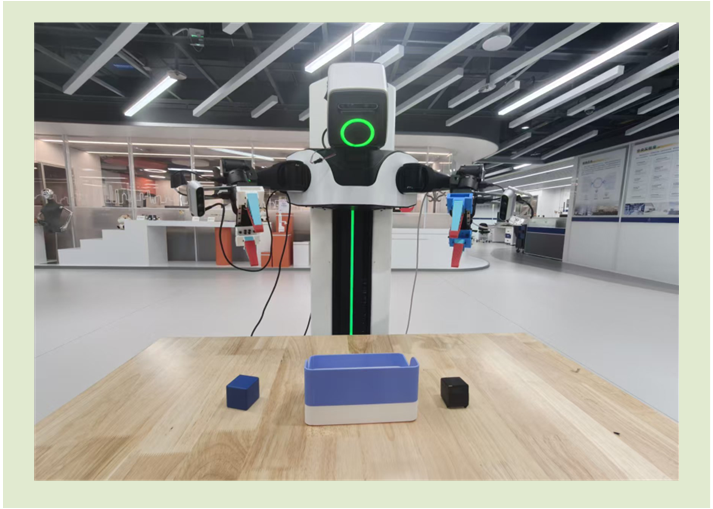}
        \vspace{-1.7em}
        \caption{``Put cubes into a box''}
        \label{fig:42}
    \end{subfigure}
    \caption{Real-world task setup. (a) shows the dual-arm handover task, while (b) illustrates the task of putting cubes into a box.}
    \label{fig:444}
\end{figure}

The real-world multi-task experiments are conducted on the AIRBOT MMK2 platform using two bimanual tasks: ``Hand over the bottle'' and ``Put cubes into a box.'' As shown in Fig.~\ref{fig:444}, the robot is equipped with three Intel RealSense D435i cameras (one head-mounted and two wrist-mounted), providing rich and complementary visual feedback for bimanual manipulation. The robot uses 3D-printed TPU grippers. We collect 100 teleoperated trajectories (50 per task). For the real-world experiments, MoE-ACT is configured with 6 experts and uses Top-2 expert selection. For each method, a multi-task policy is trained, and performance is reported as the average success rate over 30 trials per task. All model inference is performed on an RTX 3050 GPU.

\subsubsection{Real Robot Bimanual Manipulation Performance}

\begin{figure*}[!t]
    \centering
    \begin{subfigure}{0.95\linewidth}
        \centering
        \includegraphics[width=\linewidth]{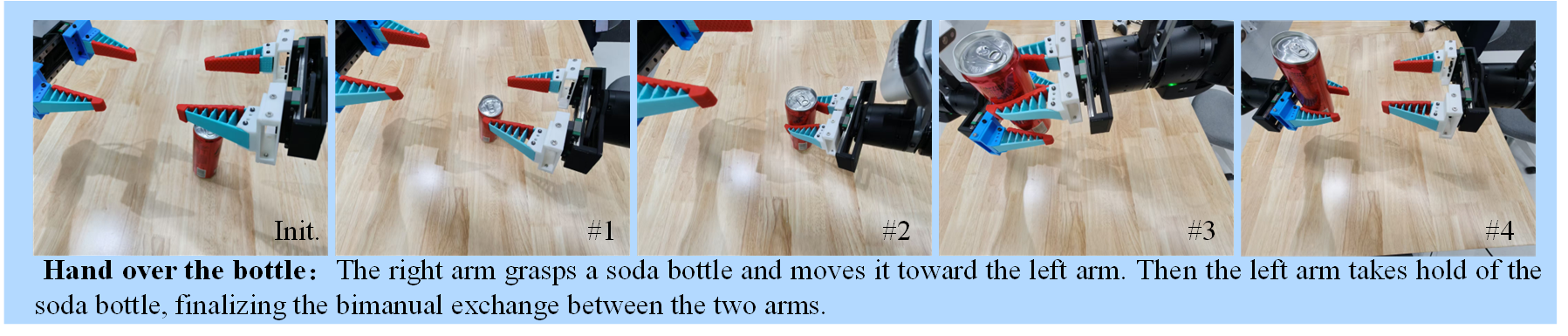}
        \vspace{-1.5em}
        \caption{Task Definition of ``Hand over the bottle'' }
        \label{fig:13}
    \end{subfigure}
    \hfill
    \begin{subfigure}{0.95\linewidth}
        \centering
        \includegraphics[width=\linewidth]{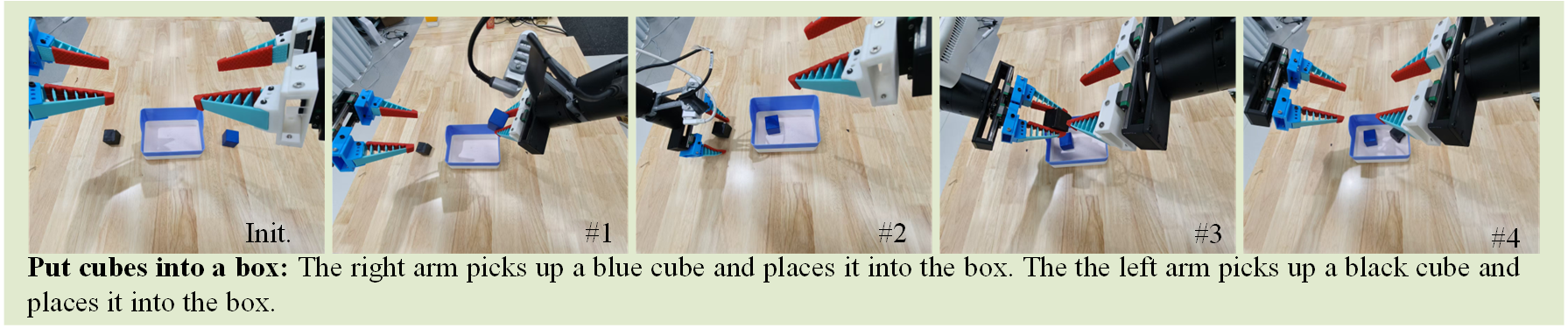}
        \vspace{-1.5em}
        \caption{Task Definition of ``Put cubes into a box''}
        \label{fig:task_cubes}
    \end{subfigure}
    \caption{Real-world task definitions. (a) shows the ``Hand over the bottle'' task, where the robot must grasp a bottle and transfer it from one arm to the other. (b) shows the ``Put cubes into a box'' task, which requires the robot to pick up cubes and place them into a box. These tasks evaluate the multi-task learning capability of MoE-ACT in real-world bimanual manipulation scenarios.}
    \label{fig:14}
\end{figure*}

As illustrated in Fig.~\ref{fig:14}, both tasks involve reaching and grasping but impose different coordination requirements: bottle handover requires coordinated grasping and release between the arms, whereas cube placement requires each arm to grasp and transport an object to a common destination. As shown in Table~\ref{tab:real_robot}, MoE-ACT achieves the highest success rate on both real-world tasks, with an average of 60.0\%, outperforming ACT and BAKU by 25.0 and 13.3 percentage points, respectively. It achieves success rates of 66.7\% on ``Hand over the bottle'' and 53.3\% on ``Put cubes into a box.'' During task execution, MoE-ACT achieves more accurate gripper positioning during grasping, particularly in grasp location and approach height. The performance gains on both tasks further support its ability to accommodate different coordination requirements within a unified policy. Overall, these results validate the effectiveness of the MoE-ACT framework for real-world multi-task bimanual manipulation.

\begin{table}[H]
\centering
\caption{Real-world experimental comparison for multi-task bimanual manipulation. Success rates (\%) are reported over 30 trials per task.}
\label{tab:real_robot}
\setlength{\tabcolsep}{8pt} 
\begin{tabular}{l|ccc}
\toprule
Methods & \makecell{Hand over \\ the bottle} & \makecell{Put cubes \\ into a box}& \makecell{Avg.}\\
\midrule
ACT\cite{2} & 43.3 & 26.7 & 35.0 \\
BAKU\cite{23} & 56.7 & 36.7 & 46.7 \\
MoE-ACT (ours) & \textbf{66.7} & \textbf{53.3} & \textbf{60.0}\\
\bottomrule
\end{tabular}
\end{table}

 \subsubsection{Failure Analysis}
 It is worth noting that the success rate of ``Put cubes into a box'' remains relatively modest. Most failures occur during grasping due to the small cube size. Small errors in grasp location or approach height can prevent stable contact, making this task particularly sensitive to gripper--object alignment. We also observe frequent failures when initial object positions are significantly shifted. A likely reason is the limited dataset size and insufficient diversity in object placements, which restrict robustness to complex scene variation. Although MoE-ACT provides capacity for task-specific feature learning through sparse expert routing, limited demonstration diversity may constrain how effectively this capacity is utilized. The MoE architecture can be expanded to support learning from more diverse manipulation data while activating only a subset of experts for each input. Future work aims to expand demonstration coverage through additional data collection and augmentation, and explore the scalability of MoE-ACT with larger, more diverse datasets.

\section{Conclusion}
This paper presents MoE-ACT, a framework for multi-task learning for dual-arm robots. By integrating sparse MoE modules into the Transformer encoder, we effectively address performance challenges in multi-task robotic manipulation. Experimental results show that MoE-ACT outperforms ACT by 17.4 percentage points in average success rate across 16 manipulation tasks on RoboTwin 2.0. Compared with $\pi_0$, MoE-ACT achieves an average success rate that is 5.1 percentage points higher. The framework also improves multi-task manipulation performance in real-world experiments. These findings highlight the value of accommodating task-specific manipulation requirements within a unified policy. At present, optimal architectures and hyperparameters still require empirical tuning. Future work will evaluate the framework on a broader range of real-world tasks. We will also explore jointly scaling expert capacity and demonstration diversity for more general-purpose manipulation, and investigate unsupervised skill discovery within the MoE routing space.

\section*{Acknowledgment}
Generative AI tools (OpenAI ChatGPT) were used for language editing and error checking. All technical content, methodology, experiments, analysis, and conclusions were developed and verified by the authors.

\bibliographystyle{IEEEtran}
\bibliography{main}
\end{document}